%% file: main.tex
\documentclass[10pt,twocolumn,letterpaper]{article}

\usepackage{cvpr}              
\usepackage{multicol}
\usepackage{multirow}
\usepackage{float}
\usepackage{subcaption}

\input{preamble}

\definecolor{cvprblue}{rgb}{0.21,0.49,0.74}
\usepackage[pagebackref,breaklinks,colorlinks,citecolor=cvprblue]{hyperref}

\def\paperID{***} 
\def\confName{CVPR}
\def\confYear{2024}

\title{\vspace{-1.7cm}UFineBench: Towards Text-based Person Retrieval with Ultra-fine Granularity}

\author{Jialong Zuo$^{1}$\quad
		 Hanyu Zhou$^1$\quad
          Ying Nie$^2$\quad
          Feng Zhang$^1$\quad
          Tianyu Guo$^2$\quad
          Nong Sang$^1$\quad\\
		Yunhe Wang$^{2*}$\quad
		Changxin Gao$^{1}$\thanks{Corresponding Authors: Changxin Gao (\href{mailto:cgao@hust.edu.cn}{cgao@hust.edu.cn}), Yunhe Wang (\href{mailto:yunhe.wang@huawei.com}{yunhe.wang@huawei.com})}\quad \\
		$^1$National Key Laboratory of Multispectral Information Intelligent Processing Technology, \\School of Artificial Intelligence and Automation, Huazhong University of Science and Technology 
        \\ $^2$Huawei Noah’s Ark Lab  \\
	}

\begin{document}
\maketitle
\begin{abstract}

Existing text-based person retrieval datasets often have relatively coarse-grained text annotations. This hinders the model to comprehend the fine-grained semantics of query texts in real scenarios. To address this problem, we contribute a new benchmark named \textbf{UFineBench} for text-based person retrieval with ultra-fine granularity.

Firstly, we construct a new \textbf{dataset} named UFine6926. We collect a large number of person images and manually annotate each image with two detailed textual descriptions, averaging 80.8 words each. The average word count is three to four times that of the previous datasets. In addition of standard in-domain evaluation, we also propose a special \textbf{evaluation paradigm} more representative of real scenarios. It contains a new evaluation set with cross domains, cross textual granularity and cross textual styles, named UFine3C, and a new evaluation metric for accurately measuring retrieval ability, named mean Similarity Distribution (mSD). Moreover, we propose CFAM, a more efficient \textbf{algorithm} especially designed for text-based person retrieval with ultra fine-grained texts. It achieves fine granularity mining by adopting a shared cross-modal granularity decoder and hard negative match mechanism.

With standard in-domain evaluation, CFAM establishes competitive performance across various datasets, especially on our ultra fine-grained UFine6926. Furthermore, by evaluating on UFine3C, we demonstrate that training on our UFine6926 significantly improves generalization to real scenarios compared with other coarse-grained datasets. The dataset and code will be made publicly available at \url{https://github.com/Zplusdragon/UFineBench}.




\end{abstract}

\vspace{-5mm}
\section{Introduction}
\begin{figure*}[htb]
\centering
\includegraphics[width=0.95\linewidth]{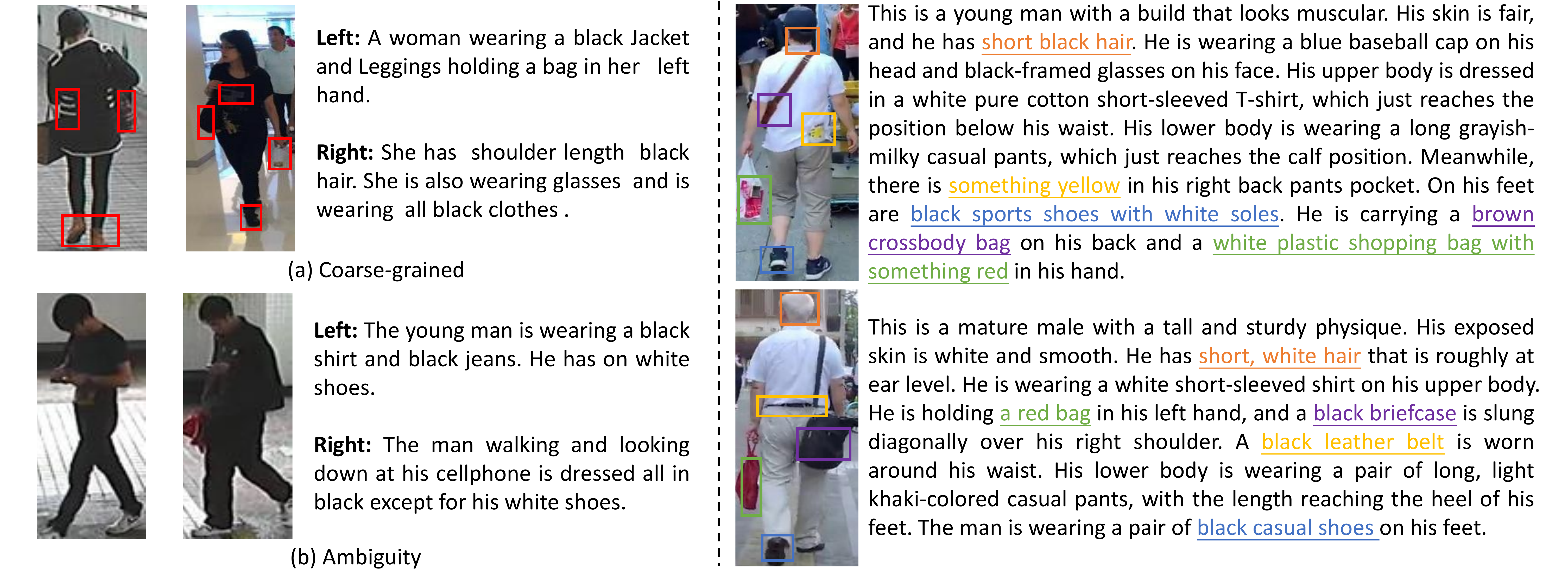}
\caption{Comparisons between our proposed UFine6926 and existing other datasets. (a)-(b) are the examples from CUHK-PEDES~\cite{cuhkpedes}. In (a), some fine-grained features not described in the text are highlighted in red boxes. In (b), the text does not provide enough details to closely match its intended identity but effectively describes the other identity. Meanwhile, two examples from UFine6926 are presented on the right, with ultra fine-grained texts. As the text details some fine-grained features in the images (highlighted in different colors correspondingly), it not only provides rich cross-modal information but also effectively distinguishes highly similar image samples.
}
\vspace{-5mm}
\label{fig:introduction}
\end{figure*}


Existing text-based person retrieval benchmarks~\cite{cuhkpedes,icfgpedes,rstpreid}, even if claimed to be fine-grained, often have coarse-grained text annotations in practice. This makes them degenerated into attribute-based retrieval~\cite{AttributeBased5,AttributeBased6,AttributeBased7,AttributeBased8} due to the provided coarse-grained descriptions to some extent. Considering this, we propose a benchmark named \textit{UFineBench} for text-based person retrieval with ultra-fine granularity, which is more in line with real scenarios.


Our work is motivated by three main aspects. As the first aspect, existing datasets~\cite{cuhkpedes,icfgpedes,rstpreid} suffer from a common issue that the text is not fine-grained enough to effectively apply to real scenarios. Specifically, as shown in Figure~\ref{fig:introduction} (a), they almost only briefly describe the common appearance of persons, and lack further specific descriptions of the unique appearance. This can easily lead to the model only being able to identify typical attribute characteristics and cannot understand the fine-grained semantics of complex query texts in real scenarios. Meanwhile, as shown in Figure~\ref{fig:introduction} (b), they suffer from the ambiguity of one identity-binding text corresponding to multiple different identities, hindering the model from accurately understanding how texts and images match during training. The detailed explanations can be found in Section~\ref{sec3.1}.

As the second aspect, existing standard evaluation sets~\cite{cuhkpedes,icfgpedes,rstpreid} all have fixed domain, fixed textual granularity and fixed textual styles. However, in real scenarios, there are usually three common features. 1) Extensive time and location coverage of surveillance videos, leading to substantial domain variations within the image gallery; 2) Inconsistency of granularity within the query texts, resulting from the variability in the actual information available for the person being searched; 3) The language expression of each describer has an unique style, even when conveying the same meaning. However, existing evaluation sets with fixed settings are inadequate for effectively assessing the model's performance in real scenarios with these features.

As the third aspect, existing evaluation metrics~\cite{cuhkpedes,IRRA} are not accurate enough to measure the retrieval ability. Given a text query, all images in the gallery are ranked according to their similarities with the query. The commonly used rank-k metric is calculated according to whether any image of the corresponding person is retrieved among the top k images. However, this calculation method of discretizing continuous similarity values leads to inaccurate measurement. For example, for the same rank conditions, the similarity conditions are highly likely to be different, but the rank-k metric cannot measure such differences.

Considering the above three aspects, this paper makes the following contributions. The first contribution is the build of a high quality dataset with ultra-fine granularity for text-based person retrieval, named \textit{UFine6926}. It contains 6,926 identities, 26,206 images and 52,412 textual descriptions. A total of 58 annotators participated in crafting the textual descriptions. Each annotator is required to provide a description according to the person's appearance as detailed as possible. Compared to existing datasets~\cite{cuhkpedes,icfgpedes,rstpreid}, the UFine6926 dataset has significant superiority in terms of textual granularity. As shown in Figure~\ref{fig:introduction}, the level of detail in each textual description has significantly improved. The average word count is 80.8 and three to four times that of the previous datasets.

The second contribution is the construction of a special evaluation set with cross domains, cross textual granularity and cross textual styles, named \textit{UFine3C}, which is more representative of real scenarios. It is collected from the test sets of the coarse-grained CUHK-PEDES~\cite{cuhkpedes}, the medium-grained ICFG-PEDES~\cite{icfgpedes} and our fine-grained UFine6926 to contain different domains, textual granularity and textual styles. Meanwhile, we utilize the large language models Qwen-14B~\cite{Qwen} and Llama2-70B~\cite{llama2} to further enrich the variations of textual granularity and styles. It contains 7,446 images to be searched and 37,939 text queries of 2,250 persons in total. 

As the third contribution, a more accurate evaluation metric is proposed for measuring retrieval ability, named mean Similarity Distribution (mSD). It is based on the continuous similarity values rather than the discrete rank conditions~\cite{cuhkpedes,IRRA,market1501,dukemtmc}. It requires the model to distinguish as much as possible the similarity differences between text queries and positive-negative image samples within a more precise numerical range. For the same rank conditions with different similarity conditions, it can sensitively measure the differences among them, while other metrics cannot. 

Based on the proposed cross-modal shared granularity decoder and hard negative match mechanism, we also contribute a novel Cross-modal Fine-grained Aligning and Matching framework (CFAM). It establishes competitive performance on various datasets without bells and whistles, especially on our fine-grained UFine6926.

\section{Related Work}
CUHK-PEDES~\cite{cuhkpedes} is the first benchmark focusing on text-based person retrieval. It contains 40,206 images and 80,412 texts of 13,003 identities. The average word count per text is 23.5. To provide a baseline algorithm, the authors propose GNA-RNN which introduces the gated neural attention mechanism into an recurrent neural network. 

However, the texts in CUHK-PEDES contain identity-irrelevant details. To address this issue, the ICFG-PEDES benchmark~\cite{icfgpedes} is constructed. There are 54,522 person images, 54,522 texts of 4,102 identities gathered from MSMT17~\cite{MSMT17}. The average word count per text is is 37.2. As a baseline algorithm, the authors propose SSAN to implement semantically self-alignment and part-level feature automatic extraction.

Meanwhile, RSTPReid~\cite{rstpreid} is constructed. It contains 20,505 images and 51,010 textual descriptions of 4,101 persons totally. The average word count per text is is 26.5. As a baseline algorithm, the authors propose DSSL which takes surroundings-person separation, fusion mechanism and five alignment paradigms into a unified framework. 

In short, these existing benchmarks are all suffering from coarse textual granularity. Therefore, it is necessary for us to propose a benchmark with ultra-fine granularity.

\section{Benchmark}
\subsection{Granularity Matters}\label{sec3.1}
Granularity-related research has become a hot topic in the computer vision field~\cite{aligntell,chan,filip,loupe,BLIP,BLIP2,nie2023lightclip,li2022transformer,he2024species196}. However, when directing to text-based person retrieval, researchers often confine themselves to a few coarse-grained benchmarks~\cite{cuhkpedes,icfgpedes,rstpreid}, thereby overlooking the significance of granularity in practical applications. We believe that the coarseness of textual granularity in existing benchmarks can give rise to the following two issues.

On the one hand, a substantial amount of coarse-grained descriptions greatly degrade the task into attribute-based retrieval~\cite{AttributeBased1,AttributeBased2,AttributeBased3,AttributeBased4}. A simple example is illustrated in Figure~\ref{fig:introduction} (a). Since the text does not describe such fine-grained features highlighted in red boxes, the model can not understand what brown boots, white stripes, and so on, refer to. When facing real scenarios with highly detailed text queries, the models trained on such coarse-grained data often prove inadequate. Meanwhile, searching for images based on coarse-grained attributes is what attribute-based person retrieval excel at. Therefore, coarse textual granularity makes these two tasks being fundamentally equivalent. 


On the other hand, coarse textual granularity introduces significant ambiguity into the training process and undermines the model's performance. As a standard practice, the optimization objectives are based on the premise that each text is only associated with the images of one identity. However, in the existing benchmarks~\cite{cuhkpedes,icfgpedes,rstpreid}, it is common that one text can be used to describe the images from different identities. A simple example is illustrated in Figure~\ref{fig:introduction} (b). Due to the overly coarseness, the text of each images cannot be highly correlated to its respective identity. Instead, it describes the other identity quite well. This ambiguity significantly hinder the model from accurately understanding how texts and images match during training.

Consequently, we emphasize that the text granularity matters and is a non-ignorable factor for text-based person retrieval. Motivated by this, we propose this benchmark with ultra-fine granularity in textual descriptions.

\subsection{Dataset with Ultra-fine Granularity}
We construct the first high quality dataset with ultra-fine granularity for this task, named \textbf{UFine6926}. It contains 26,206 images and 52,412 descriptions of 6,926 persons totally. The construction process is described as two steps:

First, while the person images in existing datasets are mostly derived from fixed-scene videos captured by stationary cameras, our dataset leverages a vast collection of unrestricted scene videos from the internet to obtain these images. We utilize the FairMOT algorithm~\cite{fairmot} to extract person tracklets from the scene videos provided by~\cite{LUPnl}. One person tracklet is considered as one identity. Then, we utilize the noise-filtering strategies proposed in PLIP~\cite{plip} to perform preliminary denoising on the obtained images. Finally, we conduct meticulous manual selection to ensure the image quality. Through this procedure, we have collected 26,206 high quality images of 6,926 identities in total.

Second, to obtain the ultra fine-grained textual descriptions, we hire 58 unique workers involved in the annotation task, instructing them to describe all important characteristics in the given images as detailed as possible. There are a total of 8,475 unique words in our dataset. Each person image is annotated with two textual descriptions. The longest description has 218 words and the average word count is 80.8, which is significantly larger than the 23.5 words of CUHK-PEDES~\cite{cuhkpedes}, 37.2 words of ICFG-PEDES~\cite{icfgpedes} and 26.5 words of RSTPReid~\cite{rstpreid}. As demonstrated by the examples in Figure~\ref{fig:introduction} and the specific statistics provided in Table~\ref{tab:compare}, our dataset exhibits a significant advantage in terms of textual granularity when compared to existing datasets.
\begin{table}
  \centering
  \resizebox{\linewidth}{!}{
  \begin{tabular}{c|c|c|c|c}
    \hline
    Dataset & Maximum &Minimum& Average&Unique\\
    \hline
    CUHK-PEDES & 96&15&23.5&9408 \\
    ICFG-PEDES & 83&9&37.2&5790 \\
    RSTPReid & 70&11&26.5&3138 \\
    
    \hline
    
    \textbf{UFine6926} & \textbf{218}&\textbf{30}&\textbf{80.8}&\textbf{8475} \\
    \hline
  \end{tabular}}
  \caption{Some statistics of texts in existing datasets. The text granularity of ours far exceeds that of others.}
  \label{tab:compare}
  \vspace{-5mm}
\end{table}

In conclusion, the properties of our UFine6926 dataset can be summarized as follows: ultra fine-grained and unfixed scene. It can be served as a benchmark to facilitate further development in this research field.

\subsection{Evaluation Set with Cross Settings}
To better evaluate the model performance in real scenarios, we construct a evaluation set named \textbf{UFine3C} with cross domains, cross textual granularity and cross textual styles based on two existing datasets~\cite{cuhkpedes,icfgpedes} and our UFine6926. This evaluation set is very challenging and the construction process is described as two steps:

First, we collect the images and textual descriptions of according persons from the test sets of the coarse-grained ``CUHK-PEDES"~\cite{cuhkpedes}, the medium-grained ``ICFG-PEDES"~\cite{icfgpedes} and our fine-grained ``UFine6926". We collect 750 persons from each of them to avoid bias and ensure fairness. After this collection, we obtain a set with spanning domains, textual granularity and textual styles.

Second, as large language models~\cite{llama,T5,PALM,GPT3,InstructGPT,vicuna,alpaca,gpt4,ChatGPT} show remarkable ability in natural language processing, we utilize Qwen-14B~\cite{Qwen} and Llama2-70B~\cite{llama2} to further enrich the variations of textual granularity and styles. Given an original description, we ask the models to response with the prompt instruction: ``\textit{Please reorganize the description in a different way. You can write it as long or as short as you like}: [\textit{original description}]". Meanwhile, we manually revise the responses generated by them to avoid incorrect answers. Through this approach, we obtain more textual descriptions with different styles and granularity.

UFine3C contains 7,446 images, 37,939 text queries of 2,250 persons totally. This evaluation set with cross settings is more consistent with real scenarios and can be served as a standard evaluation set to facilitate relevant researches. 

\subsection{A New Evaluation Metric}
\begin{figure}[htb]
\centering
\includegraphics[width=0.8\linewidth]{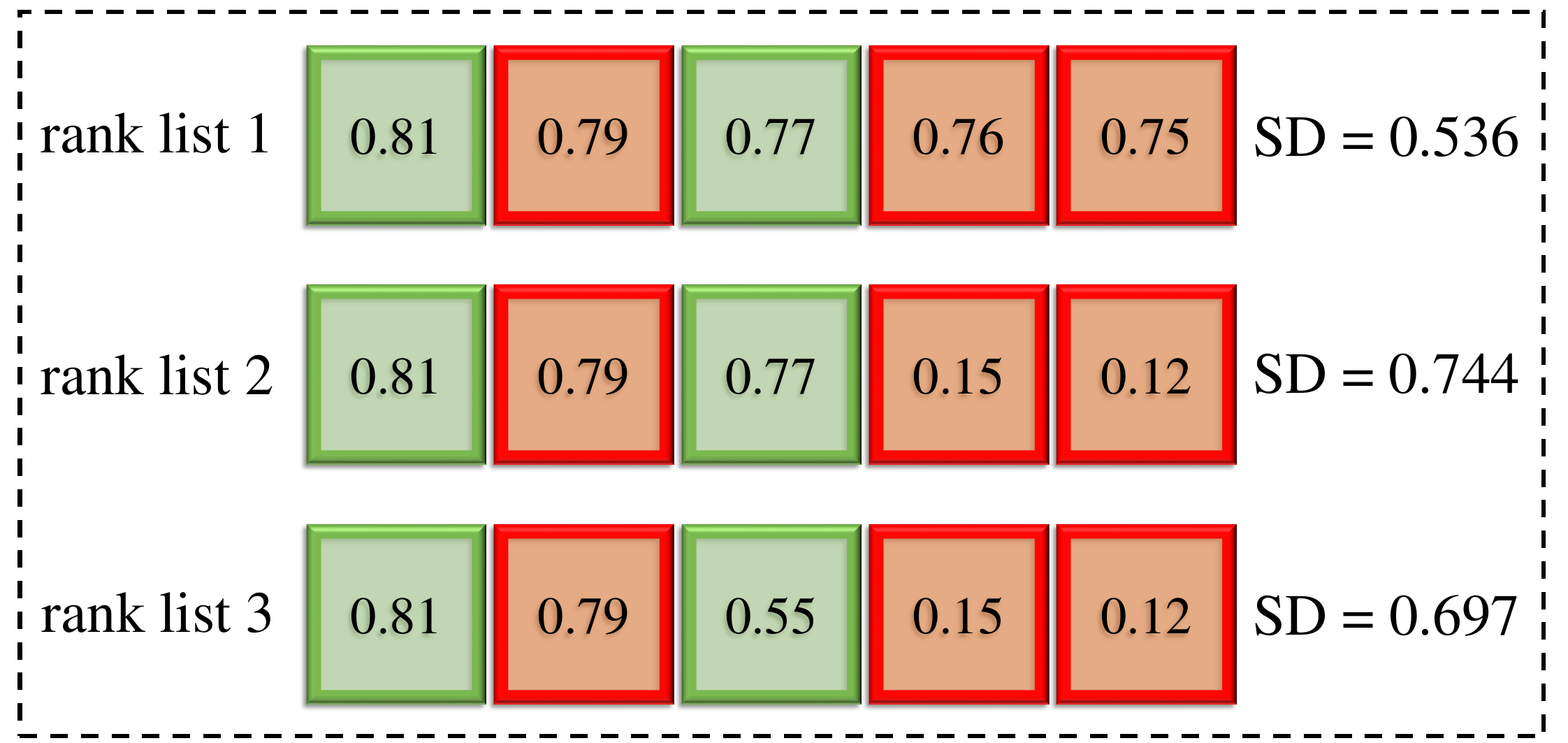}
\caption{A toy example of the difference between SD and AP metrics. Green and red boxes mean true and false matches, respectively. For these three rank lists, the AP remains 0.833. But SD = 0.536, 0.744 and 0.697, respectively.}
\label{fig：mapweak}

\end{figure}

Current benchmarks~\cite{cuhkpedes,icfgpedes,rstpreid} typically use the mean average precision (mAP) to evaluate the overall performance of person retrieval algorithms. This evaluation metric is based on discrete rank conditions and cannot sensitively measure the differences in model performance at a continuous similarity level. However, continuous similarity values more realistically reflect the model's retrieval ability. As seen in Figure~\ref{fig：mapweak}, there is a significant difference in the actual similarity values of these three rank lists. However, the APs of them both equal to 0.833, which fail to provide a fair comparison of the quality between these three rank lists.

For UFine6926 dataset, the fine-grained retrieval ability is what we especially emphasize and any difference is a reflection of it. Therefore, we propose a new metric named mean similarity distribution (mSD) to evaluate the overall performance at a continuous similarity level. As shown in Figure~\ref{fig：mapweak}, when mSD is used, the differences between these three rank lists can be well distinguished. The SDs of them are 0.536, 0.744 and 0.697, respectively. 

Given a rank list $\{s_{i}\}_{i=1}^{n}$ with $n$ ranked samples, where $s_i$ means the similarity value of the $i$-$th$ ranked sample, which is linearly normalized to the range of 0 to 1, and $s^+$ and $s^-$ means respective matched and unmatched samples. The calculation process of this metric is as follows:

First, we calculate the normalized average similarity ratio between matched samples and unmatched samples by:
\begin{equation}\label{eq1}
    PNR = 1-e^{-kx},
\end{equation}
where $x$ is the average similarity ratio between matched and unmatched samples in a list and $k$ is set to 1 as default.

Then, we calculate the average similarity precision by:
\begin{equation}\label{eq2}
    ASP= \frac{1}{n^+}\sum_{k=1}^{n^+}\frac{\sum_{i=1}^{j_k}{s_{i}^{+}}}{\sum_{i=1}^{j_k}{s_{i}}},
\end{equation}
where $\{j_k\}_{k=1}^{n+}$ means the rankings of $n^+$ matched samples.

Then, the similarity distribution (SD) of a rank list can be measured by the product of $PNR$ and $ASP$. Finally, the mean value of SDs of all rank lists, \ie, mSD, is calculated as our evaluation metric.

\subsection{Evaluation Paradigm}
During evaluation, all images in the gallery are ranked according to their similarities with the text query. We adopt the traditional rank-$k$ accuracy and mAP, and our newly proposed mSD to evaluate the retrieval performance.

\noindent
\textbf{Standard Evaluation.} As a standard in-domain evaluation paradigm, UFine6926 is divided into two subsets for training and test. The training set contains 18,577 images and 37,154 texts of 4926 identities. The test set contains 7,629 images and 15,258 text queries of 2,000 identities. 

\noindent
\textbf{Special Evaluation.} As a special evaluation paradigm with cross settings, UFine3C is utilized as a test set for evaluating the model performance in real scenarios. The training set is the same as that of standard paradigm.

\section{Method}
\subsection{Overview}
\begin{figure}[htb]
\centering
\includegraphics[width=\linewidth]{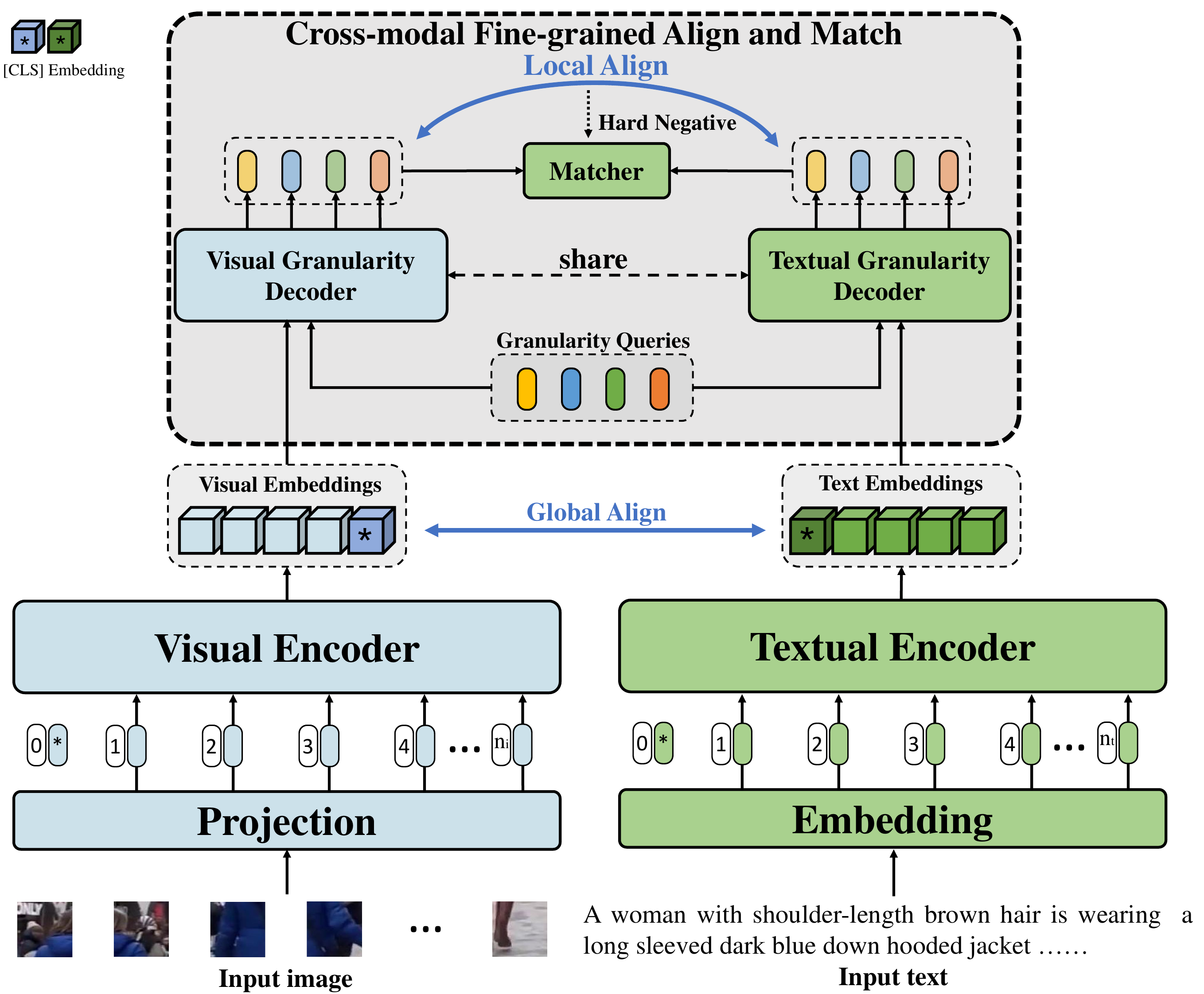}
\caption{Overview of the proposed CFAM framwork.}
\label{fig：framework}
\vspace{-5mm}
\end{figure}

In this section, we introduce a Cross-modal Fine-grained Aligning and Matching framework (CFAM), which achieves fine granularity mining in a non trivial way. The whole framework is shown in Figure~\ref{fig：framework}, given an input image $I$ and an input text $T$, the CLIP~\cite{CLIP} pre-trained visual encoder $\mathbf{E_v}$ and textual encoder $\mathbf{E_t}$ are adopted to extract the visual embeddings $\mathbf{V=\{v_1,v_2,\dots,v_{n_i}\}}$ and textual embeddings $\mathbf{W=\{w_1,w_2,\dots,w_{n_t}\}}$, respectively. Specially, we design a cross-modal fine-grained align and match module to improve the fine-grained retrieval ability. Through a shared cross-modal granularity decoder and hard negative match mechanism, the framework achieves competitive performance on various datasets and settings. 

\subsection{Cross-modal Fine-grained Aligning}
Given the extracted visual and textual embeddings, most existing methods~\cite{IRRA,plip} only calculate global similarities to achieve cross-modal alignment, which is inclined to overlooking the fine-grained details in both modalities. Therefore, we propose to perform more fine-grained alignment based on the local embeddings. However, the visual embeddings $\mathbf{V}$ and textual embeddings $\mathbf{W}$ usually have different length. To address this issue, we propose a shared cross-modal granularity decoder $\mathbf{D_g}$ with a fixed set of granularity queries $\mathbf{Q=\{q_1,q_2,\dots,q_K\}}$. These queries can interact with the embeddings and extract fine-grained information for cross-modal alignment.

For visual fine-grained information extraction, the granularity decoder $\mathbf{D_g}$ take the queries $\mathbf{Q}$ and the visual embeddings $\mathbf{V}$ as input, and then produce the fine-grained visual representations as follow,
\begin{equation}\label{eq_gv}
    \mathbf{\overline{V}=D_g(Q,V)},
\end{equation}
where $\mathbf{\overline{V}=\{\overline{v}_1,\overline{v}_2,\dots,\overline{v}_K\}}$ has the same length as the granularity queries. Meanwhile, the fine-grained textual representations $\mathbf{\overline{W}=\{\overline{w}_1,\overline{w}_2,\dots,\overline{w}_K}\}$ are produced in the similar way.

In this decoding procedure, the output representations corresponding to a certain query contain the relevant fine-grained information from both modalities and share similar semantic content. Therefore, the cross-modal similarity for each query output can be measured to achieve fine-grained alignment. We use the cosine distance to measure the similarity and the overall similarity of the $K$ query outputs can be calculated by:
\begin{equation}\label{eq_sim}
    Sim(\mathbf{V},\mathbf{W}) = \frac{1}{K} \sum_{i}^{K}{\frac{\mathbf{\overline{v}_i^\intercal\overline{w}_i}}{||\mathbf{\overline{v}_i}||||\mathbf{\overline{w}_i}||}}.
\end{equation}

 Then, given a batch of $B$ image-text pairs, the commonly used SDM loss~\cite{IRRA} will be utilized to calculate the local alignment loss $\mathcal{L}_{ls}$ according to the similarity distribution.

\subsection{Cross-modal Hard Negative Matching}
To further facilitate the cross-modal alignment, we propose to perform prediction on whether the granularity representations of each modality are matched. This task can be seen as a binary classification problem: the paired image-text is considered the positive sample, while the unpaired is considered the negative one. Unlike common random sampling, we employ the hard negative mining strategy, which is beneficial to learning more discriminative representations. 

For each image within a batch $|B|$, we sample the unpaired text whose owns the highest similarity with this image as the hard negative. Also, we sample one hard negative image for each text in the same way. Through this approach, we obtain $|B|$ positive pairs and 2$|B|$ negative pairs, denoted as $|\overline{B}|$ pairs. Then, we pass the fine-grained representations of these $|\overline{B}|$ pairs through a binary classifier named Matcher, to optimize the following objective: 
\begin{equation}\label{eq_itm}
    \mathcal{L}_m = \frac{1}{|\overline{B}|} \sum_{(\mathbf{\overline{V}},\mathbf{\overline{W}})\in \overline{B}}(\hat{y}\log{p(\mathbf{\overline{V}},\mathbf{\overline{W}})}+(1-\hat{y})(1-\log{p(\mathbf{\overline{V}},\mathbf{\overline{W}})})),
\end{equation}
where $p$ is a binary likelihood distribution function, and $\hat{y}$ is 1 if $(\mathbf{\overline{V}},\mathbf{\overline{W}})$ is matched, 0 otherwise.

\subsection{Training and Inference Strategy}

As complementary to fine-grained alignment, we compute the global similarity between the global visual embedding and the global textual embedding, and optimize the global alignment loss $\mathcal{L}_{gs}$ according to it. Also, we propose to utilize the cross-modal identity classifying loss $\mathcal{L}_{cid}$ with hard negative samples to explicitly ensure that the representations of the same image/text pair are closely clustered together. The details of this strategy are shown in the supplement. The overall training objective is the weighted sum of the above losses:

\begin{equation}\label{eq_loss}
    \mathcal{L} = \mathcal{L}_{gs} + \lambda_1\mathcal{L}_{ls} +  \lambda_2 \mathcal{L}_{m}+\lambda_3 \mathcal{L}_{cid},
\end{equation}
where $\lambda_1,\lambda_2,\lambda_3$ are hyper-parameters to adjust the weight of each loss, which are all set to 1 as the default.

In the inference phase, we will discard all additional designs and only compare the similarities between the visual and textual global embeddings. First, all images in the gallery will be passed to the visual encoder to extract the according global visual embeddings. Second, for each text query, we obtain its textual global embedding in a similar way and then we compute its similarity with the visual global embeddings of all images. Finally, we utilize the calculated similarities for ranking the image candidates.

\section{Experiments}
\subsection{Implementation}
We conduct text-based person retrieval on our proposed fine-grained UFine6926 datasets and three existing datasets CUHK-PEDES~\cite{cuhkpedes}, ICFG-PEDES~\cite{icfgpedes} and RSTPReid~\cite{rstpreid}. Meanwhile, we utilize the UFine3C evaluation set as an extra supplement to assess the generalization ability of the models in real scenarios. We adopt the popular rank-$k$ metric (k=1,5,10), the mean Average Precision (mAP) and our proposed mean Similarity Distribution (mSD) as the evaluation metrics. The higher rank-$k$, mAP and mSD indicates better performance.

CFAM mainly consists of a pre-trained visual encoder, \ie, CLIP-ViT-B/16~\cite{CLIP}, a pre-trained text encoder, \ie, CLIP textual encoder, a random-initialized granularity decoder and a matcher. The granularity decoder is shared by visual and textual modalities, consisting of 2-layer transformer blocks~\cite{transformer}. The matcher is consisted of 2-layer transformer blocks and an MLP with sigmoid activation. For each layer of the granularity decoder and matcher, the hidden size and number of heads are set to 512 and 8. The number of granularity queries is set to 16 and their hidden dimension is 512. For downstream training, the images are resized to $384\times 128$ and the maximum length of the textual tokens is set to 168. The batchsize per GPU is set to 64. Also, random erasing, horizontally flipping and crop with padding are employed for image augmentation. Random masking and replacement is employed for text augmentation. Our CFAM is trained with Adam~\cite{adam} for 60 epochs with an initial learning rate $1e^{-5}$. We adopt the linearly warm-up strategy within the beginning 5 epochs. For the random-initialized modules, the initial learning rate is set to $5e^{-5}$. We adopt the cosine learning rate decay strategy. The experiments are performed on 1 V100 32GB GPU.

\subsection{Importance of Fine Granularity}

In this section, we conduct experiments to study the importance of fine granularity in real-world scenarios. Specifically, we have trained two baseline models (PLIP~\cite{plip} and IRRA~\cite{IRRA}) and our CFAM on three coarse-grained existing datasets and our fine-grained dataset. Then, due to the fact that generalization ability is indispensable in real-world scenarios, we evaluate their performance under a range of cross settings. Please note that PLIP~\cite{plip} is a pre-trained model on a large amount of pedestrian data, giving it a SoTA generalization capability in the current field. However, compared to PLIP, CFAM still demonstrates competitive performance under a range of cross settings. 


\noindent
\textbf{Fineness Better Generalizes to Real-world Scenarios.}
In real-world scenarios, there are many variations in image domains, textual granularity and textual styles. The ability to effectively address these variations is a necessity for a high-level model. To study whether training on our proposed fine-grained UFine6926 dataset can lead the model to better generalize to the real-world scenarios than existing coarse-grained datasets~\cite{cuhkpedes,icfgpedes,rstpreid}, we conduct experiments by setting the UFine3C evaluation set as the target set to be transfered. The specific experimental procedure is described as follows. Firstly, We choose two existing popular open-source and state-of-the-art methods~\cite{plip,IRRA} and our CFAM as the baseline models. Secondly, we train the models on each training set of the three coarse-grained datasets~\cite{cuhkpedes,icfgpedes,rstpreid} and our fine-grained UFine6926. Thirdly, we directly evaluate the trained models' performance on the UFine3C evaluation set. By comparing the performance differences, we can effectively assess the generalization ability of models trained on various datasets to real-world scenarios. The experimental results are reported in Table~\ref{tab:ufine3c}. As we can see, for all the three baseline models, training on our UFine6926 dataset will significantly lead to better performance on the UFine3C evaluation set. Specifically, CFAM achieves 62.84\%, 77.82\%, 83.23\% and 46.04\% on rank-1, rank-5, rank-10 and mSD, respectively, greatly exceeding the results obtained by training on other coarse-grained datasets. We must note that the training samples in UFine6926 is much less than that in other datasets, while still achieves state-of-the-art performance. The results demonstrate that our fine-grained UFine6926 helps to learn more discriminative and general representations, which is beneficial to generalize to the real scenarios.

\noindent
\textbf{Generalization between Fineness and Coarseness.} 
We conduct the experiments under two aspects. As the first aspect, we investigate the differences in mutual generalization capabilities between coarse-grained and fine-grained datasets. We train the models on the coarse-grained datasets and then directly transfer them to our fine-grained dataset, and vice versa. The experimental results are reported in Table~\ref{tabyouliangge} (a). As we can see, for all of the three baseline models, training on the coarse-grained datasets cannot well be transferred to our fine-grained dataset. For example, when transferring to UFine6926, PLIP ~\cite{plip} trained on CUHK-PEDES~\cite{cuhkpedes} only achieves 20.52\%, 33.74\%, 42.69\% on rank-1, rank-5 and rank-10, respectively, which falls far short of practical application requirements. However, training on our fine-grained dataset can be transferred to the coarse-grained datasets to a better extent. This demonstrate that training on our fine-grained dataset enables generalization to coarse-grained datasets, while the reverse is not true. As the second aspect, we demonstrate that even when transferring to a coarse-grained dataset, training with our fine-grained dataset is mostly superior to using other coarse-grained datasets. We choose the coarse-grained ICFG-PEDES~\cite{icfgpedes} and RSTPReid~\cite{rstpreid} datasets as the target datasets. As the results reported in Table~\ref{tabyouliangge} (b), for all of the baselines, even though our dataset contains very little coarse-grained data, training on our UFine6926 still achieves better or competitive performance on the two target coarse-grained datasets. All the results demonstrate that our fine-grained dataset improves general representation learning for text-based person retrieval.

\begin{table*}[t]
\centering
\resizebox{\linewidth}{!}{
\begin{tabular}{p{2.4cm}|ccccc|ccccc||ccccc}
\hline
\multicolumn{1}{c|}{\multirow{2}{*}{Training Sets}} & \multicolumn{5}{c|}{CFAM}&\multicolumn{5}{c||}{IRRA~\cite{IRRA}}&\multicolumn{5}{c}{PLIP~\cite{plip}}\\
\cline{2-16}
 & R@1 & R@5 & R@10&mAP&mSD & R@1 & R@5 & R@10&mAP&mSD & R@1 & R@5 & R@10 &mAP&mSD \\
\hline

\multicolumn{1}{c|}{CUHK-PEDES}
&53.80 &71.05&78.25 &50.40 &38.26 
&50.06&67.98 &75.46 &47.57&36.50 
&40.45& 57.51&65.20&38.94 &30.82\\

\multicolumn{1}{c|}{ICFG-PEDES}
&36.79 &54.64&62.93&34.21 &25.47
&30.57&47.61 &55.87 &28.38&21.24  
&34.32& 50.52&57.94&32.59 &24.88\\  

\multicolumn{1}{c|}{RSTPReid}
&29.85 &49.08&58.54 &29.66 &21.82
&21.62&39.53 &49.38 &21.90&16.09  
&25.25& 40.70&48.30&24.62 &18.18 \\  

\hline

\multicolumn{1}{c|}{\textbf{UFine6926}} &\textbf{62.84} &\textbf{77.82}&\textbf{83.23} &\textbf{59.31} &\textbf{46.04} 
&\textbf{56.34}&\textbf{72.17} &\textbf{78.47}&\textbf{54.24}&\textbf{42.92}
&\textbf{64.59}&\textbf{80.16}&\textbf{85.63}&\textbf{60.43} &\textbf{47.76}\\ 
\hline
\end{tabular}}
\caption{Performance comparisons on the UFine3C evaluation dataset. The models are trained on the training sets of CUHK-PEDES, ICFG-PEDES, RSTPReid and our UFine6926, and then are directly evaluated on the UFine3C dataset. Although the training samples in UFine6926 is less than that in other datasets, training on it still achieves state-of-the-art performance. The bold results indicate the best.}
\label{tab:ufine3c}
\end{table*}

\begin{table*}
\begin{subtable}{\linewidth}
\centering
\resizebox{\linewidth}{!}{
\begin{tabular}{c|ccccc|ccccc||ccccc}
\hline
\multicolumn{1}{c|}{\multirow{2}{*}{Domains}} & \multicolumn{5}{c|}{CFAM}&\multicolumn{5}{c||}{IRRA~\cite{IRRA}}&\multicolumn{5}{c}{PLIP~\cite{plip}}\\
\cline{2-16}
 & R@1 & R@5 & R@10&mAP&mSD & R@1 & R@5 & R@10&mAP&mSD & R@1 & R@5 & R@10 &mAP&mSD \\
\hline

\multicolumn{1}{c|}{CUHK$\rightarrow$ UFine}
&42.49 &59.47&68.14 &45.06 &33.74
&37.63&54.99 &64.46 &40.79&30.80 
&20.52& 33.74&42.69&24.17 &18.90\\

\multicolumn{1}{c|}{ICFG$\rightarrow$UFine} 
&20.65 &34.66&43.05 &23.09 &16.65 
&14.99&26.85 &33.92 &17.02&12.29 
&12.13& 21.88&28.73&14.98 &11.20\\

\multicolumn{1}{c|}{RSTP$\rightarrow$UFine}
&20.20 &35.31&44.02 &23.13 &16.73
&13.13&25.59 &33.81 &15.55&11.20
&9.75& 18.86&25.27&12.32 &8.95\\

\hline
\multicolumn{1}{c|}{UFine$\rightarrow$CUHK}
&48.72 &70.21&78.17 &44.42 &33.23
&41.41&62.72 &71.85 &39.22&29.86 
&56.53& 77.24&84.10&51.60 &39.85 \\

\multicolumn{1}{c|}{UFine$\rightarrow$ICFG}
&40.78 &60.90&69.31 &22.30 &16.28
&35.08&55.16 &64.02 &18.87&13.85 
&51.52& 70.96&78.02&27.67 &20.93 \\ 

\multicolumn{1}{c|}{UFine$\rightarrow$RSTP} 
&45.10 &72.35&81.45 &35.40 &25.40
&41.30&64.25 &76.00 &32.04&22.93 
&43.85& 72.10&80.60&33.88 &24.97\\ 

\hline
\end{tabular}}
\caption{Differences in mutual generalization capabilities between coarse-grained and fine-grained datasets.}
\vspace{2mm}
\end{subtable}

\begin{subtable}{\linewidth}
\resizebox{\linewidth}{!}{
\begin{tabular}{c|ccccc|ccccc||ccccc}
\hline
\multicolumn{1}{c|}{\multirow{2}{*}{Domains}} & \multicolumn{5}{c|}{CFAM}&\multicolumn{5}{c||}{IRRA~\cite{IRRA}}&\multicolumn{5}{c}{PLIP~\cite{plip}}\\
\cline{2-16}
 & R@1 & R@5 & R@10&mAP&mSD & R@1 & R@5 & R@10&mAP&mSD & R@1 & R@5 & R@10 &mAP&mSD \\
\hline

\multicolumn{1}{c|}{CUHK $\rightarrow$ ICFG} 
&46.21 &65.18&72.65 &24.77 &18.00
&42.42&62.07 &69.64 &21.80&15.94
&53.81& 72.56&79.34&30.20 &22.76\\

\multicolumn{1}{c|}{RSTP$\rightarrow$ICFG}
&38.55&55.37&63.53 &24.66 &18.46
&32.37&49.71 &57.75 &20.57&15.44 
&51.01& 69.52&76.71&32.14 &23.82 \\ 

\multicolumn{1}{c|}{UFine$\rightarrow$ICFG}
&40.78 &60.90&69.31 &22.30 &16.28 
&35.08&55.16 &64.02 &18.87&13.85 
&51.52& 70.96&78.02&27.67 &20.93\\ 

\hline
\multicolumn{1}{c|}{ICFG$\rightarrow$CUHK}
&40.48 &63.48&72.68 &37.38 &27.11 
&33.45&56.12 &66.21 &31.39&22.78 
&56.40& 76.98&83.82&51.72 &39.12\\ 

\multicolumn{1}{c|}{RSTP$\rightarrow$CUHK}
&40.11 &63.55&72.61 &37.29 &27.04
&32.67&55.20 &65.34 &30.17&21.87
&50.15&72.84&81.01&46.84 &34.43\\ 

\multicolumn{1}{c|}{UFine$\rightarrow$CUHK}
&48.72 &70.21&78.17 &44.42 &33.23
&41.41&62.72 &71.85 &39.22&29.86 
&56.53& 77.24&84.10&51.60 &39.85\\

\hline
\end{tabular}}
\caption{Fine-grained dataset can even better be transferred to other coarse-grained datasets than the coarse-grained dataset. }
\vspace{-2mm}
\end{subtable}

\caption{Performance comparisons on the generalization performance between our fine-grained UFine6926 and three existing datasets CUHK-PEDES~\cite{cuhkpedes}, ICFG-PEDES~\cite{icfgpedes} and RSTPReid~\cite{rstpreid}. The arrow direction indicates the source dataset and the target dataset.}
\label{tabyouliangge}
\end{table*}

\begin{figure}[htb]
\centering
\includegraphics[width=\linewidth]{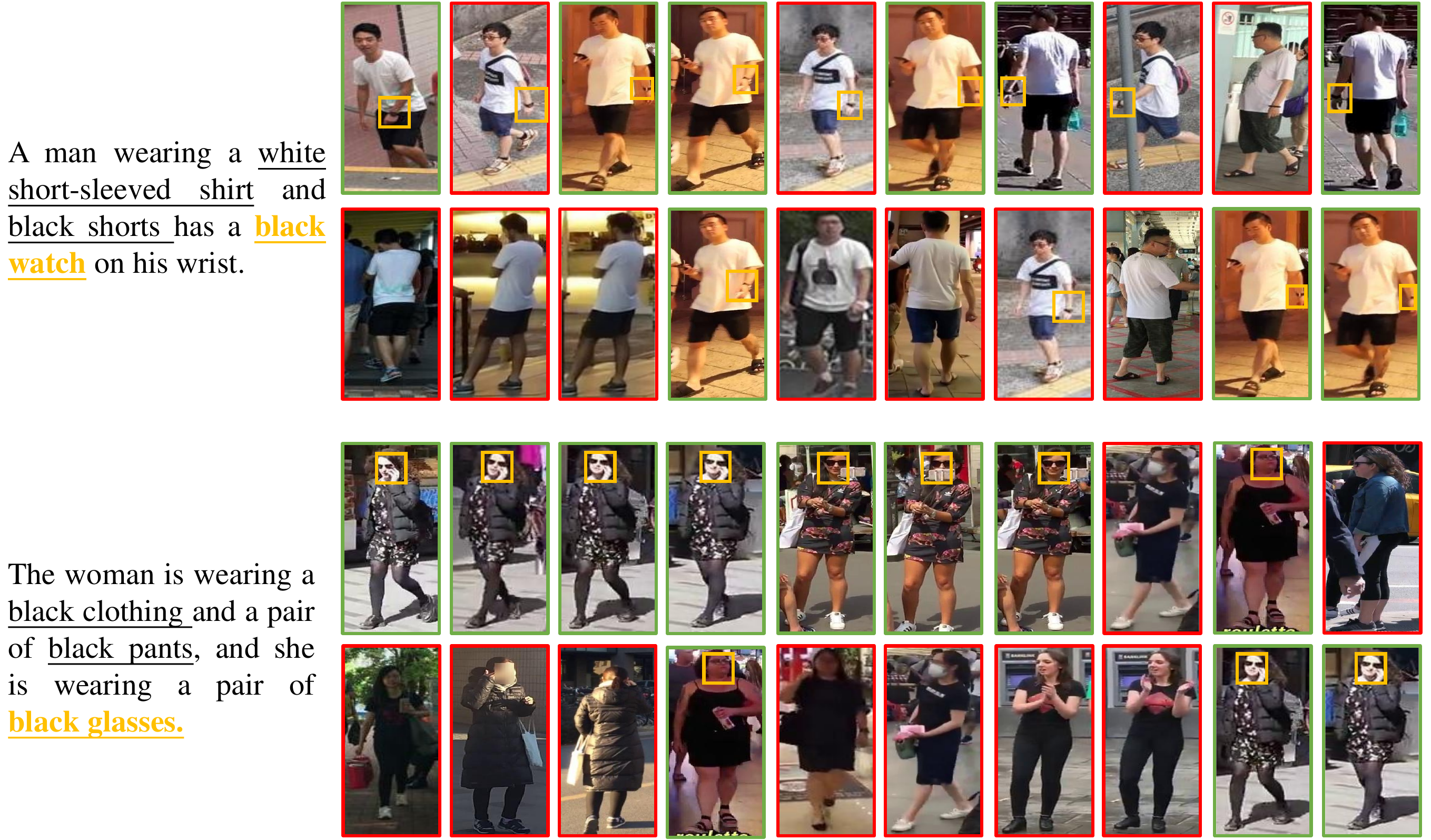}
\caption{Comparison of rank-10 retrieval results on UFine3C between CFAM trained on UFine6926~\cite{cuhkpedes} (the first row) and CUHK-PEDES (the second raw) for each textual description. The images that fully match the text are marked in green, and the unmatched ones are marked in red.}
\label{fig：qualitative}
\vspace{-5mm}
\end{figure}

\noindent
\textbf{Qualitative Results.} To make a more realistic comparison of the models' performance in real-world scenarios, we conduct a straightforward qualitative experiment. We choose our CFAM trained on the coarse-grained CUHK-PEDES~\cite{cuhkpedes} and the fine-grained UFine6926 as the baseline models to be compared. Then, we manually provide any textual descriptions to search the according persons in the UFine3C dataset. The rank-10 retrieval results from the CFAM models trained on CUHK-PEDES and UFine6926 respectively are compared in Figure~\ref{fig：qualitative}. As it shows, training on UFine6926 achieves more accurate retrieval results and can fully perceive fine-grained discriminative clues to distinguish different persons, while training on CUHK-PEDES fails to do so. This is illustrated in the orange highlighted text and image region box in Figure~\ref{fig：qualitative}. 

\subsection{Comparison with State-of-the-Art Methods}
In this section, we compare the performance of our proposed CFAM framework with state-of-the-art (SoTA) methods on our fine-grained UFine6926 dataset and three public coarse-grained datasets~\cite{cuhkpedes,icfgpedes,rstpreid}.

\noindent
\textbf{Performance Comparisons on UFine6926.}
We utilize two evaluation sets for the performance comparison on UFine6926. The first is the UFine3C evaluation set. The second is the UFine6926 test set. We evaluate the performance of existing SoTA methods trained on our new UFine6926. As the results shown in Table~\ref{ufine6926results}, on each evaluation set, CFAM outperforms all other SoTA methods. Specifically, with CLIP-ViT-L/14~\cite{CLIP} setting, it achieves 62.84\% rank-1 accuracy, 59.31\% mAP and 46.04\% mSD on UFine3C, respectively. Meanwhile, it achieves 88.51\% rank-1 accuracy, 57.09\% mAP and 68.45\% mSD on UFine6926 test set, respectively. These results demonstrate the superior fine-grained retrieval capability of CFAM. 

\begin{table}[t]
\centering
\tiny
\resizebox{\linewidth}{!}{
\begin{tabular}{c|c|ccccc}
\hline
\multicolumn{1}{c|}{\multirow{2}{*}{}} & \multicolumn{1}{c|}{\multirow{2}{*}{Methods}}&\multicolumn{5}{c}{Metrics}\\
\cline{3-7}
& & R@1& R@5& R@10 &mAP&mSD \\
\hline
\multicolumn{1}{c|}{\multirow{6}{*}{\rotatebox{90}{UFine3C}}}
&{NAFS~\cite{NAFS}} &43.69 &61.34&69.72&39.31&30.32 \\
&{LGUR~\cite{LGUR}} &51.26 &69.67&75.32&49.22&38.13\\
&{SSAN~\cite{icfgpedes}} &53.67 &71.15&77.15&51.40&39.66 \\
&{IRRA~\cite{IRRA}} &56.34 &72.17&78.47&54.24&42.92 \\
\cline{2-7}
&{CFAM(B/16)}&58.51 &74.92&80.90&55.53&43.22 \\
&{\textbf{CFAM(L/14)}}&\textbf{62.84}&\textbf{77.82}&\textbf{83.23}&\textbf{59.31}&\textbf{46.04} \\

\hline
\multicolumn{1}{c|}{\multirow{6}{*}{\rotatebox{90}{UFine6926}}}
&{NAFS~\cite{NAFS}} &64.11 &80.32&85.05&63.47&49.61 \\
&{LGUR~\cite{LGUR}} &70.69 &84.57&89.91&68.93&56.23 \\
&{SSAN~\cite{icfgpedes}} &75.09 &88.63&92.84&73.14&59.41 \\
&{IRRA~\cite{IRRA}} &83.53 &92.94&95.95&82.79&66.35 \\
\cline{2-7}
&{CFAM(B/16)} &85.55 &94.51&97.02&84.23&66.49 \\
&{\textbf{CFAM(L/14)}}&\textbf{88.51} &\textbf{95.58}&\textbf{97.49}&\textbf{87.09}&\textbf{68.45} \\

\hline
\end{tabular}}
\caption{We train some state-of-the-art open-source models on UFine6926 and evaluate the performance under two evaluation settings. We show the best score in bold .}
\label{ufine6926results}
\vspace{-3mm}
\end{table}

\noindent
\textbf{Performance Comparisons on Other Datasets.}
The experimental results on the CUHK-PEDES~\cite{cuhkpedes}, ICFG-PEDES~\cite{icfgpedes} and RSTPReid~\cite{rstpreid} datasets are reported in Table~\ref{tabchuk}, Table~\ref{tabicfg} and Table~\ref{tabRSTP}, respectively. On CUHK-PEDES, with the CLIP-ViT-B/16 setting, CFAM achieves competitive results to recent state-of-the-art methods without bells and whistles, achieving 72.87\% rank-1 accuracy, 64.92\% mAP and 50.20\% mSD, respectively. Meanwhile, with the CLIP-ViT-L/14 setting, the performance of CFAM can be further improved, achieving 75.60\% rank-1, 67.27\% mAP and 51.83\% mSD, respectively. This means that our method has good scalability. On ICFG-PEDES and RSTPReid, our CFAM also outperforms all state-of-the-art methods by a considerable margin. It achieves 65.38\% rank-1, 39.42\% mAP and 30.29\% mSD on ICFG-PEDES, 62.45\% rank-1, 49.50\% mAP and 36.92\% mSD on RSTPReid, respectively. The results demonstrate that CFAM helps to learn general representations on various datasets.

\begin{table}[t]
\tiny
\centering
\resizebox{\linewidth}{!}{
\begin{tabular}{c|ccccc}
\hline
\multicolumn{1}{c|}{\multirow{2}{*}{Method}} & \multicolumn{5}{c}{CUHK-PEDES}\\
\cline{2-6}
&   R@1 & R@5 & R@10 & mAP & mSD  \\
\hline

\multicolumn{1}{c|}{MIA~\cite{MIA}} &53.10& 75.00& 82.90&-&-\\

\multicolumn{1}{c|}{TIMAM~\cite{TIMAM}} &54.51& 77.56& 79.27&-&-\\

\multicolumn{1}{c|}{TDE~\cite{TDE}} &55.25& 77.46& 84.56&-&-\\

\multicolumn{1}{c|}{NAFS~\cite{NAFS}} &59.94& 79.86& 86.70&54.07&-\\

\multicolumn{1}{c|}{SSAN~\cite{icfgpedes}} &61.37 &80.15 &86.73&-&-\\

\multicolumn{1}{c|}{LapsCore~\cite{Lapscore}} &63.40 &- &87.80&-&-\\

\multicolumn{1}{c|}{TIPCB~\cite{tipcb}} &64.26& 83.19&89.10&-&-\\ 

\multicolumn{1}{c|}{CAIBC~\cite{CAIBC}} &64.43& 82.87&88.37&-&-\\ 

\multicolumn{1}{c|}{LGUR~\cite{LGUR}}&65.25 &83.12 &89.00&-&-\\

\multicolumn{1}{c|}{IVT~\cite{IVT}} &65.59 &83.11 &89.21&-&-\\

\multicolumn{1}{c|}{PLIP~\cite{plip}} &69.23 &85.84 &91.16&-&- \\

\multicolumn{1}{c|}{CFine~\cite{CFine}}&69.57 &85.93 &91.15&-&-\\

\multicolumn{1}{c|}{IRRA~\cite{IRRA}}&73.38 &89.93 &93.71&66.13 &51.49\\

\hline
\multicolumn{1}{c|}{CFAM(B/16)}&72.87  &88.61 &92.87	&64.92 &50.20\\
\multicolumn{1}{c|}{\textbf{CFAM(L/14)}}&\textbf{75.60}  &\textbf{90.53} &\textbf{94.36}&\textbf{67.27} &\textbf{51.83}\\

\hline
\end{tabular}}
\caption{Comparison with the state-of-the-art methods on CUHK-PEDES~\cite{cuhkpedes}. We show the best score in bold.}
\label{tabchuk}
\vspace{-5mm}
\end{table}

\begin{table}[t]
\tiny
\centering

\resizebox{\linewidth}{!}{
\begin{tabular}{c|ccccc}
\hline
\multicolumn{1}{c|}{\multirow{2}{*}{Method}} & \multicolumn{5}{c}{ICFG-PEDES}\\
\cline{2-6}
&   R@1 & R@5 & R@10 & mAP & mSD  \\
\hline
\multicolumn{1}{c|}{Dual Path~\cite{dualpath}} &38.99 &59.44 &68.41&-&-\\

\multicolumn{1}{c|}{CMPM/C~\cite{CMPM}} &43.51&65.44 &74.26&-&-\\

\multicolumn{1}{c|}{ViTAA~\cite{vitaa}} &50.98 &68.79 &75.78&-&-\\

\multicolumn{1}{c|}{SSAN~\cite{icfgpedes}} &54.23 &72.63 &79.53&-&-\\

\multicolumn{1}{c|}{LGUR~\cite{LGUR}}&59.02 &75.32 &81.56&-&-\\

\multicolumn{1}{c|}{IVT~\cite{IVT}} &56.04 &73.60 &80.22&-&-\\

\multicolumn{1}{c|}{CFine~\cite{CFine}}&60.83 &76.55 &82.42&-&-\\

\multicolumn{1}{c|}{IRRA~\cite{IRRA}} &63.46 &80.25 &85.82&38.06&29.54 \\

\multicolumn{1}{c|}{PLIP~\cite{plip}}&64.25 &80.88 &86.32&- &-\\

\hline
\multicolumn{1}{c|}{CFAM(B/16)}&62.17  &79.57&85.32	&36.34 &28.01\\
\multicolumn{1}{c|}{\textbf{CFAM(L/14)}}&\textbf{65.38}  &\textbf{81.17} &\textbf{86.35}	&\textbf{39.42} &\textbf{30.29}\\

\hline
\end{tabular}}
\caption{Comparison with the state-of-the-art methods on ICFG-PEDES~\cite{icfgpedes}. We show the best score in bold.}
\label{tabicfg}
\end{table}

\begin{table}[t]
\tiny
\centering

\resizebox{\linewidth}{!}{
\begin{tabular}{c|ccccc}
\hline
\multicolumn{1}{c|}{\multirow{2}{*}{Method}} & \multicolumn{5}{c}{RSTPReid}\\
\cline{2-6}
&   R@1 & R@5 & R@10 & mAP & mSD  \\
\hline

\multicolumn{1}{c|}{DSSL~\cite{rstpreid}} &39.05 &59.44 &68.41&-&-\\

\multicolumn{1}{c|}{SSAN~\cite{icfgpedes}} &43.50 &67.80 &77.15&-&-\\

\multicolumn{1}{c|}{LBUL~\cite{LBUL}} &45.55 &68.20 &77.85&-&-\\

\multicolumn{1}{c|}{IVT~\cite{IVT}}&46.70&70.00 &78.80&-&-\\

\multicolumn{1}{c|}{CFine~\cite{CFine}} &50.55 &72.50&81.60&-&-\\

\multicolumn{1}{c|}{IRRA~\cite{IRRA}}&60.20 &81.30 &88.20&47.17 &35.22\\

\hline
\multicolumn{1}{c|}{CFAM(B/16)}&59.40 &81.35 &88.50	&46.04 &34.27\\
\multicolumn{1}{c|}{\textbf{CFAM(L/14)}}&\textbf{62.45}  &\textbf{83.55} &\textbf{91.10}	&\textbf{49.50} &\textbf{36.92}\\
\hline
\end{tabular}}
\caption{Comparison with the state-of-the-art methods on RSTPReid~\cite{rstpreid}. We show the best score in bold.}
\label{tabRSTP}
\end{table}

\subsection{Ablation Study}
\begin{table}[t]
\large
\centering

\resizebox{\linewidth}{!}{
\begin{tabular}{c|cccc|ccccc}
\hline 
\multicolumn{1}{c|}{\multirow{2}{*}{No.}}  & \multicolumn{4}{c|}{Components}&\multicolumn{5}{c}{CUHK-PEDES}\\
\cline{2-10}
& $\mathcal{L}_{gs}$ &$\mathcal{L}_{ls}$ &$\mathcal{L}_m$& $\mathcal{L}_{vap}$ &R@1&R@5&R@10&mAP&mSD\\
\hline 
0 &&& &&68.45	&86.50&91.68&61.28&46.31	\\
1&\checkmark&& &&70.42	&87.20&92.22	&63.00&48.38	\\
2 &\checkmark&\checkmark& & &71.69	&87.87&92.37	&63.85&49.12	\\
3&\checkmark&\checkmark& \checkmark&&72.42	&88.31&92.80	&64.81&49.84	\\
\hline
4&\checkmark&\checkmark&\checkmark&\checkmark&\textbf{72.87}&\textbf{88.61}&\textbf{92.87}	&\textbf{64.92}&\textbf{50.20}	\\
\hline
\end{tabular}}
\caption{Ablation study on each component of CFAM.}
\label{ablation}
\vspace{-5mm}
\end{table}

To verify the contribution of each component in CFAM, we conduct an ablation experiment on CUHK-PEDES dataset~\cite{cuhkpedes}. The results are reported in Table~\ref{ablation}. No.0 is the baseline utilizing the original InfoNCE loss in CLIP~\cite{CLIP} to align the cross-modal global embeddings. As we can see, each component facilitates the model's capability, and combining all of them leads to the best performance. In addition, we have conducted further ablation experiments, which are detailed in the supplement.

\section{Conclusion}
This paper introduces a new benchmark for text-based person retrieval with ultra-fine granularity. We firstly contribute a manually annotated dataset named UFine6926 with ultra fine-grained texts. Meanwhile, we propose a special evaluation paradigm more representative for real scenarios with a new evaluation set named UFine3C and a new metric named mSD. Then, CFAM is proposed in the attempt to achieve fine-grained cross-modal representation correlation. Our benchmark will enable research possibilities in multiple directions, \textit{e.g.}, fine-grained retrieval, real scenario generalization, multi-granularity adaptation, efficient structure, \textit{etc}. We believe this work will shed light on more future researches in this community.

\section{Acknowledgement}
This work was supported by the National Natural Science Foundation of China No.62176097, and Hubei Provincial Natural Science Foundation of China No.2022CFA055. We gratefully acknowledge the support of \href{https://www.mindspore.cn/}{MindSpore}, CANN (Compute Architecture for Neural Networks) and Ascend AI Processor used for this research.

{
    \small
    \bibliographystyle{ieeenat_fullname}
    \bibliography{main}
}

\input{sec/X_suppl}

\end{document}

%% file: preamble.tex
\usepackage[dvipsnames]{xcolor}


%% file: sec/X_suppl.tex
\clearpage
\setcounter{page}{1}
\newpage
       \twocolumn[
        \centering
        \Large
        \textbf{UFineBench: Towards Text-based Person Retrieval with Ultra-fine Granularity}\\
        \vspace{0.5em}Supplementary Material \\
        \vspace{1.0em}
       ] 

\section*{Contents}
In this supplementary material, we will 1) show the details of our proposed cross-modal identity classifying loss $\mathcal{L}_{cid}$, 2) show more results of the ablation study, and 3) show more specific and compared examples of our proposed UFineBench and existing other datasets~\cite{cuhkpedes,icfgpedes,rstpreid}.

\section{Cross-Modal Identity Classifying}
In the training phase of CFAM, we also propose the cross-modal identity classifying loss $\mathcal{L}_{cid}$ as a supplement to explicitly ensure that the representations of the same image/text pair are closely clustered together. 

First, referring to~\cite{CMPM}, we revisit the traditional identity loss commonly used in the person re-identification task. Given the extracted embeddings $\mathcal{X} = \{\boldsymbol{x}_i\}_{i=1}^N$ and the identity labels $\mathcal{Y} = \{y_i\}_{i=1}^N$, the traditional identity loss can be computed by:
\begin{equation}
    \mathcal{L}_{id}=\frac{1}{N} \sum_{i}-\log \left(\frac{\exp \left(\boldsymbol{W}_{y_{i}}^{\top} \boldsymbol{x}_{i}+b_{y_{i}}\right)}{\sum_{j} \exp \left(\boldsymbol{W}_{j}^{\top} \boldsymbol{x}_{i}+b_{j}\right)}\right),
\end{equation}
where $\boldsymbol{W}_{y_i}$ and $\boldsymbol{W}_j$ denote the $y_i$-th and $j$-th column of classification weight matrix $\boldsymbol{W}$, $y_i$ indicates the identity label of $\boldsymbol{x}_i$, and $b_{y_i}$ and $b_j$ represent the $y_i$-th and $j$-th element of bias vector.

However, this loss only performs identity clustering on individual modalities, lacking interaction across different modalities and unable to conduct feature clustering in a shared cross-modal space. Therefore, we propose the cross-modal identity classifying loss, which not only considers the cross-modal correlation but also deeply mines hard negative unmatched sample pairs.

Given the extracted visual embeddings $\mathcal{V} = \{\boldsymbol{v}_i\}_{i=1}^N$ and textual embeddings $\mathcal{T} = \{\boldsymbol{t}_i\}_{i=1}^N$, for each $\boldsymbol{v}_i$ within $\mathcal{V}$, we sample the unpaired textual embedding which owns the highest similarity with this $\boldsymbol{v}_i$ as the negative. Also, we sample one hard negative visual embedding for each $\boldsymbol{t}_i$ within $\mathcal{T}$ in the same way. Specially, we add an extra identity label as the unmatched label, that is, if the original identity labels are from $M$ classes, the $(M+1)$-th identity will be set as the unmatched label. Through this approach, we obtain $|N|$ original positive pairs with original identity labels and 2$|N|$ negative pairs with an unmatched label, denoted as $|B|$ pairs with the identity labels $\{y_i\}_{i=1}^B$.

Then, for the $|B|$ pairs $\{\boldsymbol{v}_i,\boldsymbol{t}_i,y_i\}_{i=1}^B$, we first concatenate the visual embedding $\boldsymbol{v}_i$ and textual embedding $\boldsymbol{t}_i$ to form the cross-modal embedding $\boldsymbol{z}_i$. Then, for $\{\boldsymbol{z}_i\}_{i=1}^B$ with the identity labels $\{y_i\}_{i=1}^B$, the cross-modal identity classifying loss can be computed by:
\begin{equation}
    \mathcal{L}_{cid}=\frac{1}{N} \sum_{i}-\log \left(\frac{\exp \left(\boldsymbol{W}_{y_{i}}^{\top} mlp(\boldsymbol{z}_{i})+b_{y_{i}}\right)}{\sum_{j} \exp \left(\boldsymbol{W}_{j}^{\top} mlp(\boldsymbol{z}_{i})+b_{j}\right)}\right),
\end{equation}
where $mlp$ denotes an MLP layer consisted of a Linear layer, a LayerNorm, a GELU activation to fuse the cross-modal embeddings more deeply.

\section{More Results of Ablation Study}
\begin{table}[t]
\large
\centering

\resizebox{\linewidth}{!}{
\begin{tabular}{c|ccc|ccccc}
\hline 
\multicolumn{1}{c|}{\multirow{2}{*}{No.}}  & \multicolumn{3}{c|}{Components}&\multicolumn{5}{c}{CUHK-PEDES}\\
\cline{2-9}
& share &depth &queries& R@1&R@5&R@10&mAP&mSD\\
\hline 
0 &&1& 16&71.17	&87.83&92.72&63.83&49.00	\\
1 &&2&16 &71.05	&87.56&92.58&63.79&49.10	\\
2 &&3&16 &70.96	&87.69&92.31&63.56&48.79	\\
3 &&4&16 &71.72	&87.76&92.75&64.13&49.13	\\

4 &\textbf{\checkmark}&\textbf{2}&\textbf{4} &\textbf{72.87}	&\textbf{88.61}&92.87&\textbf{64.92}&\textbf{50.20}\\

5 &\checkmark&2&8 &71.72	&88.34&93.00&64.32&49.64\\
6 &\checkmark&2&12 &71.61	&88.30&92.72&64.27&49.72\\
7 &\checkmark&2&16 &71.83	&88.43&93.15&64.39&49.52\\
8 &\checkmark&2&20 &72.08	&88.48&93.02&64.50&49.51\\

9 &\checkmark&3&4 &72.09	&88.48&93.05&64.44&49.68\\
10 &\checkmark&3&8 &71.59	&88.61&\textbf{93.34}&64.06&49.06\\
11 &\checkmark&3&12 &72.34	&88.50&93.00&64.40&49.38\\
12 &\checkmark&3&16 &71.85	&88.32&92.95&64.23&49.40\\
13 &\checkmark&3&20 &72.24	&88.55&82.92&64.40&49.64\\

14 &\checkmark&4&4 &71.41	&88.30&93.02&64.17&49.73\\
15 &\checkmark&4&8 &72.13	&88.56&93.02&64.32&49.35\\
16 &\checkmark&4&12 &71.61	&88.32&92.98&64.37&49.63\\
17 &\checkmark&4&16 &72.04	&88.58&92.97&64.44&49.66\\
18 &\checkmark&4&20 &72.06	&88.42&93.00&64.41&49.80\\
\hline
\end{tabular}}
\caption{Ablation study on some components of CFAM. The ``share" denotes whether the granularity decoder is shared across modalities. The ``depth" denotes the number of transformer blocks in the granularity decoder. The ``queries" represents the number of query tokens used to extract fine-grained information.}
\label{ablation}
\vspace{-5mm}
\end{table}
We conduct an ablation experiment to study the influence of whether the granularity decoder is shared across modalities, the number of transformer blocks in the granularity decoder and the number of query tokens. The models are trained and evaluated on the CUHK-PEDES dataset~\cite{cuhkpedes}. According to the results in the Table~\ref{ablation}, we can draw two conclusions as follows. First, compared to an unshared granularity decoder (No.1, No.2, No.3), a shared granularity decoder (No.7, No12, No.17) can bring about a significant improvement in performance. Second, when the number of transformer blocks in the granularity decoder and query tokens are 2 and 16 (No.4), respectively, the best performance is obtained, achieving 72.87\% rank-1, 64.92\% mAP, and 50.20\% mSD, which is set as the default in CFAM.

\begin{figure*}[htb]
\centering
\includegraphics[width=\linewidth]{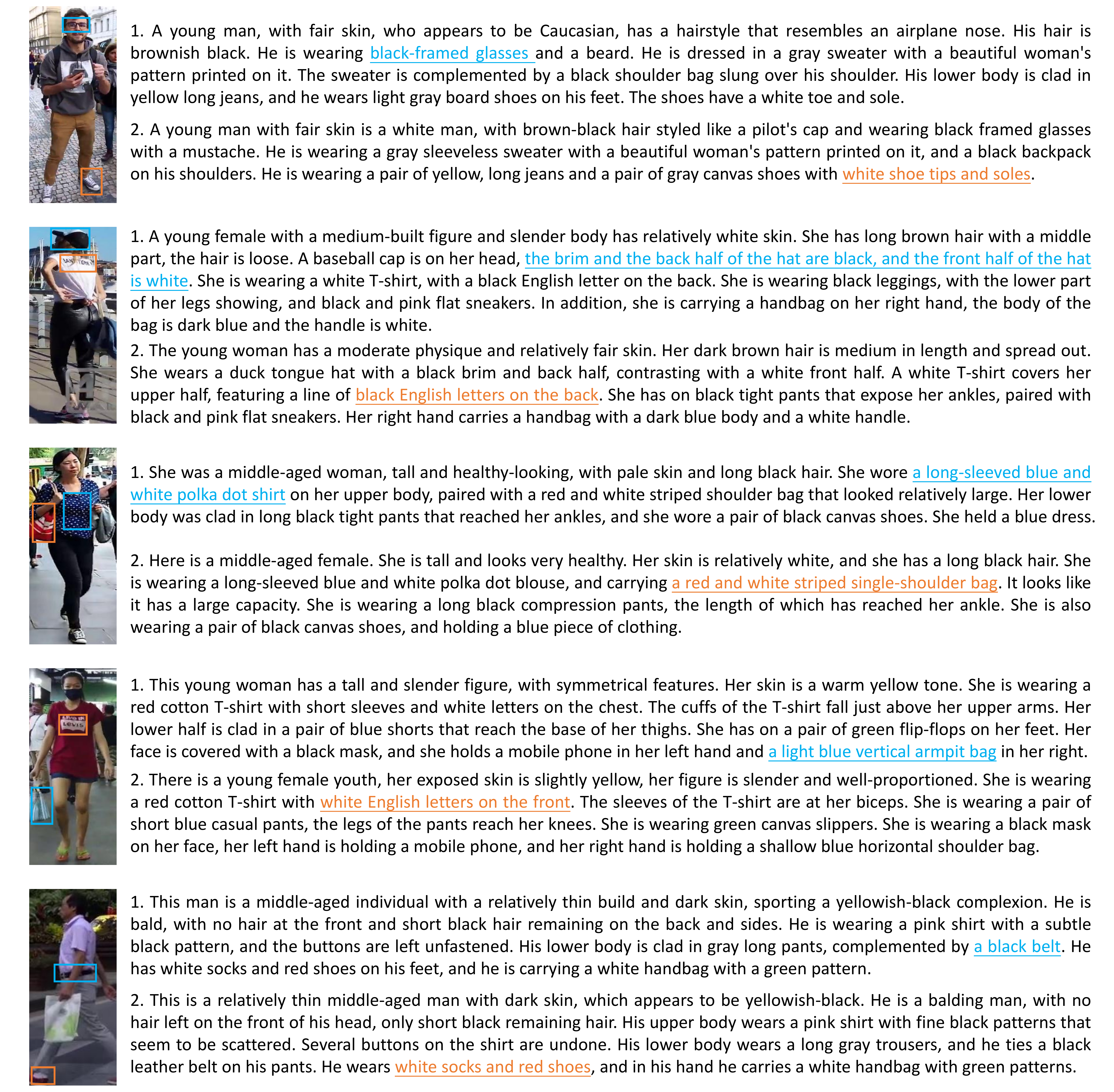}
\caption{Some examples of our proposed UFine6926. Every image has two different fine-grained textual descriptions that describes the person's apperance detailedly. Some fine-grained features are highlighted in blue or orange boxes and texts accordingly.}
\label{fig：ufine6926}
\vspace{-5mm}
\end{figure*}

\begin{figure*}[htb]
\centering
\includegraphics[width=\linewidth]{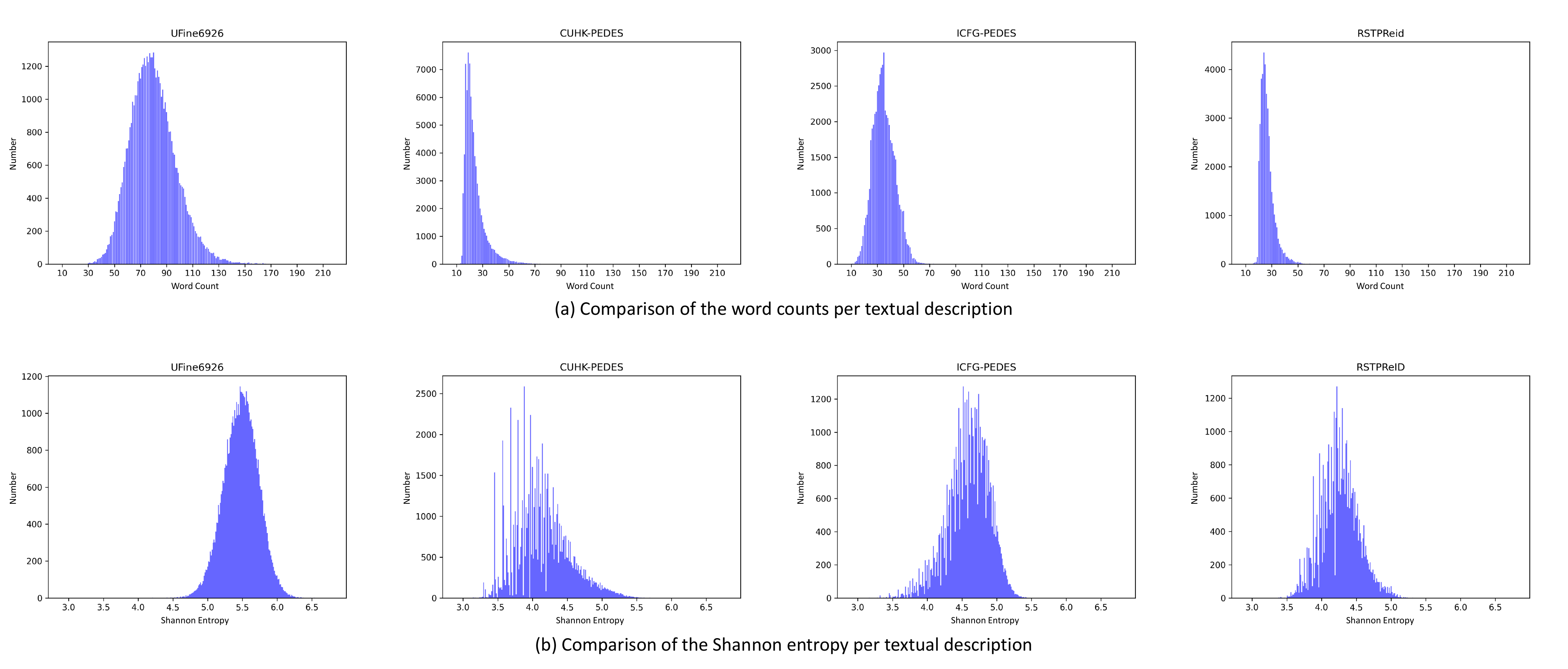}
\caption{Statistical comparison of the word counts per textual description in our UFine6926 with those in other datasets~\cite{cuhkpedes,icfgpedes,rstpreid}.}
\label{fig：statistic}
\end{figure*}

\section{More Examples of UFineBench}
\noindent
\textbf{Ultra Fine-grained UFine6926.}
We have shown more representative examples of our proposed ultra fine-grained UFine6926 in Figure~\ref{fig：ufine6926}. As we can see, each person image is annotated with two different textual descriptions that describes the person's appearance detailedly. Even if the external characteristics of certain persons in the images are very subtle, our textual descriptions do not ignore them like the previous datasets~\cite{cuhkpedes,icfgpedes,rstpreid} and portrays these fine-grained features accurately. These fine-grained features are highlighted in blue or orange boxes and texts in the Figure~\ref{fig：ufine6926}. For instance, in the bottom example, the area of the person's shoes is very inconspicuous, yet our textual description accurately identifies this area and describes it as ``white socks and red shoes." 
Meanwhile, we conduct a statistical comparison of the word counts per textual description in our UFine6926 with those in other datasets, as illustrated in Figure~\ref{fig：statistic} (a). By examining the distribution, we can observe that the word count for UFine6926 is generally centered around 80, while simultaneously, CUHK-PEDES~\cite{cuhkpedes} and RSTPReid~\cite{rstpreid} are both roughly centered around 25, and ICFG-PEDES~\cite{icfgpedes} is centered around 30. It is evident that the level of textual detail in our UFine6926 is higher than that in all other datasets by a significant margin.
At the same time, in information theory, Shannon entropy is often used to measure the amount of information in a system. For a text, we can treat each word as an event, calculate the probability of each word occurring, and finally calculate the entropy of the entire text based on Shannon's entropy formula. We conduct a statistical comparison of the Shannon entropy per textual description in our UFine6926 with those in other datasets, as illustrated in Figure~\ref{fig：statistic} (b). Meanwhile, we also calculate the specific Shannon entropy metrics for each dataset, as shown in Table~\ref{shannon_metrics}. The higher the entropy, the greater the amount of information in the text. From the distribution and metrics, we can see that our UFine6926 has significantly richer textual information than other datasets.  

\begin{table}[t]
\large
\centering
\resizebox{\linewidth}{!}{
\begin{tabular}{ccccc}
\hline 
Benchmark & MaxEnt & MinEnt & AverEnt & AllEnt \\
\hline 
CUHK-PEDES & 5.901 & 3.057 & 4.140 & 6.956 \\
ICFG-PEDES & 5.604 & 3.122 & 4.593 & 6.412 \\
RSTPReid & 5.496 & 2.914 & 4.265 & 6.448 \\
\hline
\textbf{UFine6926} & \textbf{6.621} & \textbf{4.201} & \textbf{5.480} & \textbf{7.465} \\
\hline
\end{tabular}}
\caption{Comparison of entropy-based metrics for each dataset. ``MaxEnt", ``MinEnt", ``AverEnt" represent text-level maximum, minimum and average entropy, respectively. ``AllEnt" is the dataset-level entropy calculated by aggregating all the texts in the dataset. The higher the entropy, the richer the information.
}
\label{shannon_metrics}
\vspace{-5mm}
\end{table}

\begin{figure*}[htb]
\centering
\includegraphics[width=\linewidth]{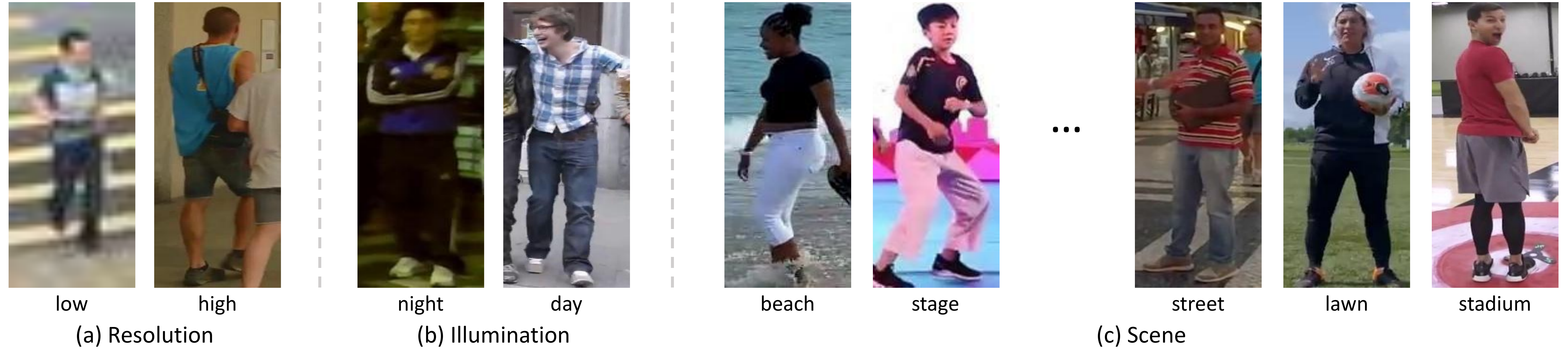}
\caption{Our UFine3C spans across various \textbf{domains} such as image resolution, illumination, and shooting scenes.}
\label{fig:ufine3c1}
\end{figure*}

\begin{figure*}[htb]
\centering
\includegraphics[width=\linewidth]{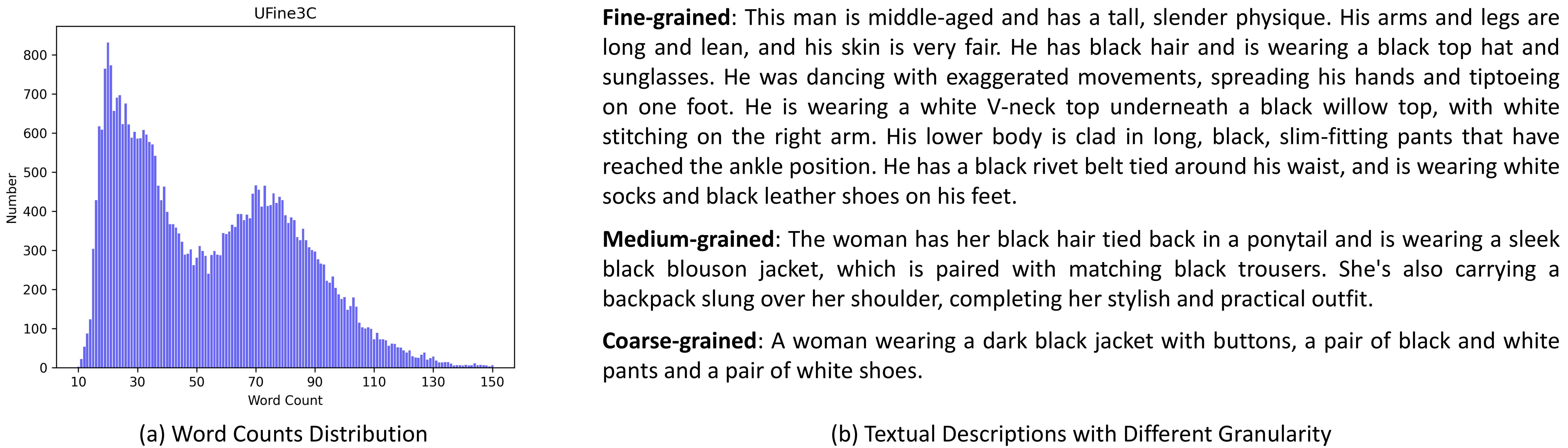}
\caption{Our UFine3C spans across various \textbf{text granularity} from coarse-grained to fine-grained. The word counts distribution is shown in (a). The (b) illustrates some specific examples representing three different text granularity conditions.}
\label{fig:ufine3c2}
\end{figure*}

\begin{figure*}[htb]
\centering
\includegraphics[width=\linewidth]{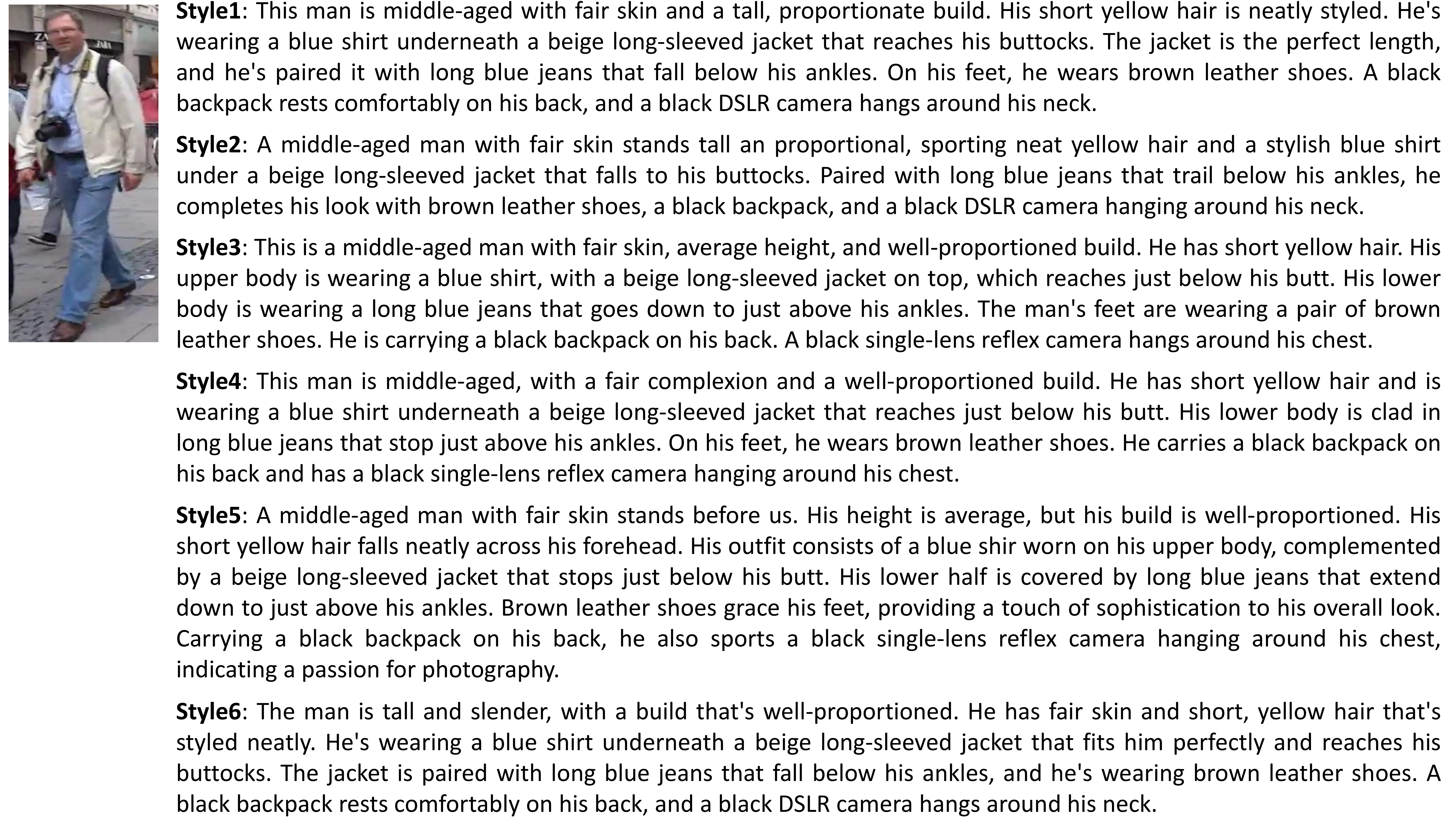}
\caption{Our UFine3C spans across various \textbf{text styles}, simulating the language expression styles of different individuals.}
\label{fig:ufine3c3}
\end{figure*}

\noindent
\textbf{Three Cross Settings in UFine3C.}
Our proposed UFine3C evaluation set has cross domains, cross text granularity and cross text styles, which is more representative of the challenges faced in real scenarios. (1) \textit{Our UFine3C spans across various domains.} As shown in Figure~\ref{fig:ufine3c1}, the images in UFine3C have significant variations in resolution, illumination and shooting scenes, which is very close to the situation of images obtained in real scenarios. (2) \textit{Our UFine3C spans across various text granularity.} As shown in Figure~\ref{fig:ufine3c2}, on the left, the word counts distribution of UFine3C is spanning from coarse-grained to fine-grained. Meanwhile, on the right, according to the text granularity, the texts can be categorized into the fine-grained, the medium-grained and the coarse-grained, respectively. This represents the inconsistency of granularity within the query texts in real scenarios. (3) \textit{Our UFine3C spans across various text styles.} As shown in Figure~\ref{fig:ufine3c3}, each image has multiple query texts with different styles, simulating the language expression styles of different individuals in practice. 

